# Heads or Tails: A Simple Example of Causal Abstractive Simulation

Gabriel Simmons

gsimmons@ucdavis.edu

**Abstract**

This note illustrates how a variety of causal abstraction (Beckers & Halpern, 2019; Rubenstein et al., 2017), defined here as *causal abstractive simulation*, can be used to formalize a simple example of language model simulation. This note considers the case of simulating a fair coin toss with a language model. Examples are presented illustrating the ways language models can fail to simulate, and a success case is presented, illustrating how this formalism may be used to prove that a language model simulates some other system, given a causal description of the system. This note may be of interest to three groups. For practitioners in the growing field of language model simulation, causal abstractive simulation is a means to connect ad-hoc statistical benchmarking practices to the solid formal foundation of causality. Philosophers of AI and philosophers of mind may be interested as causal abstractive simulation gives a precise operationalization to the idea that language models are *role-playing* (Shanahan, 2024). Mathematicians and others working on causal abstraction may be interested to see a new application of the core ideas that yields a new variation of causal abstraction.

## 1 Background

To keep the emphasis on the example, I will list some background assumptions without argument, but with some references to supporting literature.

1. The relations of *implementation* and *abstraction* are relations between pairs of systems (or kinds of systems). Both relations require causal consistency between the systems. See Chalmers (2011); Chalmers (1996) for implementation, and (Rubenstein et al., 2017), (Beckers et al., 2019; Beckers & Halpern, 2019) for abstraction.

2. The relation of *simulation* is like the relation of implementation and abstraction in that it requires causal consistency. The differences between simulation and implementation have been argued in (Dennett, 1978; Dreyfus, 1978; Pattee, 1989; Searle, 1980; Webb, 1991).

3. For practical purposes, simulation is observer-relative. In other words, System $X$ can simulate System $Y$ for Person $A$ but not for Person $B$. This view is supported, at least for the case of language model simulation, by the fact that researchers disagree about whether language models can or do simulate certain systems. See, for example, Argyle et al. (2023), Aher et al. (2023), Park et al. (2023) as examples positing simulation and Schröder et al. (2025) arguing against simulation. The need for a formal model of these disagreements motivates this note. The idea that simulation requires some involvement from an observer has likely been argued by many; it is captured particularly clearly by Webb (1991).

4. Simulations consist of Referents, Observers, Simulators, and Simulacra.

    1. *Referents* are the entities or kinds of entities that are being simulated. Referents may be real or abstract.
    2. *Observers* observe and interact with the simulation. Observers may be interested in assessing the quality of simulacra.
    3. *Simulators* are the entities tasked with simulating referents, producing simulacra. Kinds of entities often used as simulators include computers and language models.
    4. *Simulacra* are likenesses that are the products of simulations. A simulation is successful when the joint properties and behaviors of the observer and the simulator are such that a likeness of the referent appears for the observer; this apparent likeness is the simulacrum.

This terminology has been used with somewhat similar meanings by Baudrillard (1994).

5. The observer's model of a system is a causal model (Pearl, 2009; Spirtes et al., 2000). There is evidence in support of the view that causal models offer a good approximation of human causal reasoning (Gopnik et al., 2004; Rehder, 2003; Sloman, 2005). Humans routinely violate some of the key assumptions of formal causal inference (Rehder & Burnett, 2005; Waldmann et al., 2008), but amendments to the formal models are available (Davis & Rehder, 2020; Rehder, 2024).

6. Language models may be usefully viewed as simulators. By now, this has been argued by many. I first encountered the idea in janus (2022).

7. Causal abstraction is a mathematical formalism that aims to capture what it means for one system to be an abstraction of another (Beckers et al., 2019; Beckers & Halpern, 2019; Rubenstein et al., 2017). Systems are represented by causal models, and the relation of abstraction is defined in terms of commutativity of a diagram consisting of the two models and mappings between their states. The ideas of causal abstraction will be used here to define a relation of *causal abstractive simulation*.

## 2 Causal Models

This section defines causal models and interventions. The notation here is taken, with slight modifications, from Beckers & Halpern (2019).

**Definition 1. Causal Model**
A causal model $M$ is a tuple $(\mathcal{U}, \mathcal{V}, \mathcal{R}, \mathcal{F}, \mathcal{I})$, consisting of a set of exogenous variables $\mathcal{U}$, a set of endogenous variables $\mathcal{V}$, and a function $\mathcal{R}$ that maps each variable to its range of possible values. $\mathcal{I}$ is a set of allowed interventions. Functions $\mathcal{F} = \{F_{X_i} : X_i \in \mathcal{V}\}$ are structural equations. ∎

In general, each structural equation has signature $F_{X_i} : \mathcal{R}(\mathcal{U}) \times \mathcal{R}(\mathcal{V} - \{X_i\}) \to \mathcal{R}(X_i)$ for each endogenous variable $X_i \in \mathcal{V}$. When the causal model is acyclic, we can choose an ordering of the variables such that the graph is topologically sorted, and we can write the structural equations as $F_{X_i} : \mathcal{R}(\mathcal{U}) \times \mathcal{R}(\mathcal{V}_{0..i-1}) \to \mathcal{R}(X_i)$.

**Definition 2. Probabilistic Causal Model**
A probabilistic causal model $M$ is a tuple $(\mathcal{U}, \mathcal{V}, \mathcal{R}, \mathcal{F}, \mathcal{I}, \Pr)$. A probabilistic causal model is identical to a causal model (Definition 1), with the addition of a probability distribution $\Pr$ over contexts, where a context is a setting of the exogenous variables. $\Pr$ assigns a likelihood in $[0, 1]$ to each context. ∎

"Settings of variables" will be used interchangeably with "states" in this note.

## 3 Causal Abstractive Simulation

In addition to the background assumptions in Section 1, we will assume the following:

**Proposition 1. Causal Abstractive Simulation Hypothesis (CASH)**
A simulator $L$ simulates a referent $R$ to an observer $O$ if $O$'s model of $R$ is a causal abstraction of $L$. ∎

The point of this note is not to argue for this hypothesis, but to illustrate how it can be operationalized. Arguments for the CASH and its connections to the causal abstraction literature will appear in future work. Much credit is due to Beckers et al. (2019); Beckers & Halpern (2019); Rubenstein et al. (2017) for the development of the causal abstraction formalism, on which the CASH relies heavily.

## 3.1 Referents

The referent's true causal structure is represented by a causal model $M_R$. Neither the observer nor the simulator has direct access to $M_R$.

**Definition 3. Referent**
A referent $R = \langle M_R \rangle$ where $M_R$ is a causal model $M_R = \langle \mathcal{U}_R, \mathcal{V}_R, \mathcal{R}_R, \mathcal{F}_R, \mathcal{I}_R \rangle$. ∎

## 3.2 Simulators

A simulator is a probabilistic causal model $M_L$.

**Definition 4. Simulator**
A simulator $L = \langle M_L \rangle$ where $M_L$ is a probabilistic causal model $M_L = \langle \mathcal{U}_L, \mathcal{V}_L, \mathcal{R}_L, \mathcal{F}_L, \mathcal{I}_L, \Pr_{\mathcal{U}_L^I} \rangle$. ∎

It is assumed that all values $\mathcal{V}_L$ are legible to observers who interact with the simulator. We assume that the exogenous variables $\mathcal{U}_L$ can be partitioned into two sets: $\mathcal{U}_L = \mathcal{U}_L^O \cup \mathcal{U}_L^I, \mathcal{U}_L^O \cap \mathcal{U}_L^I = \emptyset$ where $\mathcal{U}_L^O$ are the exogenous variables that the observer can set, and $\mathcal{U}_L^I$ are exogenous variables that the observer cannot set. The observer can set the exogenous variables $\mathcal{U}_L^O$ to any value in $\mathcal{R}_L(\mathcal{U}_L^O)$. A distribution over the exogenous variables that the observer cannot set ($\Pr_{\mathcal{U}_L^I}$) is included in the definition of the simulator.

**Proposition 2. Independence Assumption**
For these examples, we assume independence between the values of $\mathcal{U}_L^O$ and $\mathcal{U}_L^I$. The observer cannot see the values of $\mathcal{U}_L^I$ to make a correlated choice of $\mathcal{U}_L^O$, and the simulator has no knowledge of $\mathcal{U}_L^O$ to make a correlated choice of $\mathcal{U}_L^I$. The case where $\mathcal{U}_L^O$ is not set is undefined; intuitively, nothing happens – the simulator is "off". ∎

## 3.3 The Observer

The observer has a probabilistic causal model of the referent $M_O$. The observer interacts with the simulator by setting the values of the exogenous variables of the simulator, and perceiving the values of the endogenous variables of the simulator. It is up to the observer to interpret these variable values in terms of the referent, in other words to map states of the simulator to states of the referent. This interpretation via state mapping is constitutive of the observer's "seeing" a simulacrum of the referent.

**Definition 5. Observer**

Given a referent $R = \langle M_R \rangle$, an observer (interested in a simulation of $R$) is defined as $O = \langle M_O, \Pr_O, \Pr_{\mathcal{U}_L \mid \mathcal{U}_O, \mathcal{I}_O}, \tau_O \rangle$, where

- $M_O = \langle \mathcal{U}_O, \mathcal{V}_O, \mathcal{R}_O, \mathcal{F}_O, \mathcal{I}_O \rangle$ is a causal model of the referent as imagined or inferred by the observer.
- $\Pr_{\mathcal{U}_O}$ is a probability distribution over contexts for the referent as imagined or inferred by the observer.
- $\Pr_{\mathcal{I}_O \mid \mathcal{U}_O}$ is a probability distribution over interventions on the referent, conditional on referent states.
- $\Pr_{\mathcal{U}_L \mid \mathcal{U}_O, \mathcal{I}_O} : \mathcal{R}_L(\mathcal{U}_L) \times \mathcal{R}_O(\mathcal{U}_O) \times \mathcal{I}_O \to [0,1]$ is a conditional probability distribution over low-level exogenous variable values (inputs to the simulator) given high-level states (exogenous variable settings of the referent) and high-level interventions (interventions on the referent). This distribution describes the likelihoods that the observer represents a given high-level state to the simulator in a particular way.
- $\tau_O : \mathcal{R}_L(\mathcal{V}_L) \to \mathcal{R}_O(\mathcal{V}_O)$ is a function that maps low-level endogenous variable settings to high-level endogenous variable settings.

∎

In practice, observers may be single humans (either scientists or non-scientists), or the behavior of "the observer" may be enacted by a group of humans. For example, a group of researchers may decide collectively on a specification of $M_O$. Additionally, human observers may choose to delegate some of their interpretative work to other systems (computer programs, AIs, etc). Going further, nothing in this formalism prevents the observers from being entirely non-human entities.

### 3.4 Causal Abstractive Simulation

#### 3.4.1 The Simulator's Inputs

The observer has a model of the referent that includes a probability distribution over contexts $\Pr_{\mathcal{U}_O}$ and a probability distribution over interventions $\Pr_{\mathcal{I}_O \mid \mathcal{U}_O}$. Once the referent is in a particular post-interventional state (in the observer's mental model), the observer translates this post-interventional state into inputs to the simulator via the distribution $\Pr_{\mathcal{U}_L \mid \mathcal{U}_O, \mathcal{I}_O}$. Thus, the distribution over observer-accessible simulation inputs is given by the following marginalization:

$$\Pr_{\mathcal{U}_L^O} = \sum_{\mathcal{U}_O} \sum_{\mathcal{I}_O} \Pr_{\mathcal{U}_L^O \mid \mathcal{U}_O, \mathcal{I}_O} \Pr_{\mathcal{I}_O \mid \mathcal{U}_O} \Pr_{\mathcal{U}_O} \tag{1}$$

Given the independence assumed between $\mathcal{U}_L^O$ and $\mathcal{U}_L^I$ (see Proposition 2), the distribution over all exogenous variables of the simulator is:

$$\Pr_L(\mathcal{U}_L^O = u^O, \mathcal{U}_L^I = u^I) = \Pr_{\mathcal{U}_L^O}(\mathcal{U}_L^O = u^O) \times \Pr_{\mathcal{U}_L^I}(\mathcal{U}_L^I = u^I) \tag{2}$$

for all settings of $\mathcal{U}_L^O$ and $\mathcal{U}_L^I$.

**Definition 6. Causal Abstractive Simulation**
Given an observer $O = \langle M_O, \text{Pr}_O, \text{Pr}_{\mathcal{J}_O \mid \mathcal{U}_O}, \text{Pr}_{\mathcal{U}_L \mid \mathcal{U}_O, \mathcal{J}_O}, \tau_O \rangle$, a simulator $L = \langle M_L, \text{Pr}_L \rangle$ is a causal abstractive simulation of $R$ for $O$ if

$$M_O(\text{Pr}_O) = \tau_O(M_L(\text{Pr}_L)) \tag{3}$$

∎

The left- and right-hand sides of the equation in Definition 6 are probability distributions over endogenous variables of the referent. $M_O(\text{Pr}_O)$ is the distribution over endogenous variable settings of the referent obtained from the structural equations and context distribution of the referent model. $M_L(\text{Pr}_L)$ is the distribution over endogenous variable settings of the simulator obtained from the structural equations and context distribution of the simulator's causal. $\tau_O(M_L(\text{Pr}_L))$ is the distribution over endogenous states of the referent obtained by mapping each endogenous state of the simulator to a referent state. This equation in Definition 6 expresses a commutativity condition between the observer's model of the referent and the behavior of the simulator under the observer's state mappings.

As noted by Beckers et al. (2019), in many cases such a commutativity condition is not satisfied exactly. Instead, we can say that the simulator is an approximate causal abstractive simulation of the referent if the difference between distributions is small:

**Definition 7. Approximate Causal Abstractive Simulation**
Given an observer $O = \langle M_O, \text{Pr}_O, \text{Pr}_{\mathcal{J}_O \mid \mathcal{U}_O}, \text{Pr}_{\mathcal{U}_L \mid \mathcal{U}_O, \mathcal{J}_O}, \tau_O \rangle$, and a simulator $L = \langle M_L, \text{Pr}_L \rangle$, $L$ is an approximate causal abstractive simulation of $R$ for $O$ if

$$d\Big( M_O(\text{Pr}_O), \tau_O\big(M_L(\text{Pr}_L)\big) \Big) < \varepsilon \tag{4}$$

given some choice of $\varepsilon > 0$ and distance $d$ between probability distributions.

∎

## 4 The Observer's Model of the Coin

The observer's model of the coin consists of causal model $M_O = \langle \mathcal{U}_O, \mathcal{V}_O, \mathcal{R}_O, \mathcal{F}_O, \mathcal{J}_O \rangle$ and probability distribution over contexts $\text{Pr}_O$. Coins are so widely used in probability texts that readers may immediately associate a coin with pure randomness. But, in fact, a coin toss is a causal system. Once I toss a coin, the face on which it will land is determined by the laws of physics, and causally influenced by factors like coin tosser hand position, the orientation of the coin, and so on. We can gloss over these details and say that prior to being tossed, the coin can be in either a heads-causing or tails-causing state. Let us represent this by exogenous variable $S$.

$$\mathcal{U}_O = \{S\}, \mathcal{R}_O(S) = \{H\text{-causing}, T\text{-causing}\} \tag{5}$$

The endogenous variable $X$ captures how the coin actually lands.

$$\mathcal{V}_O = \{X\}, \mathcal{R}_O(X) = \{H, T\} \tag{6}$$

The structural equation for this simple model of a coin captures our inuition that coins land on heads from heads-causing states and on tails from tails-causing states:

$$\mathcal{F}_O = \{F\}, F = \{H\text{-causing} \to H, \\ T\text{-causing} \to T\} \tag{7}$$

Human observers think of coins as "random" because we have a convention of tossing coins in ways that prevent us from distinguishing heads-causing and tails-causing states. We represent this by specifying $\Pr_{\mathcal{U}_O}$ as the uniform distribution:

$$\Pr_{\mathcal{U}_O}(S) = \{H\text{-causing} \to 0.5, \\ T\text{-causing} \to 0.5\} \tag{8}$$

A coin tosser can defy convention, and place the coin in a controlled fashion with a desired side facing up. We model this as an intervention where $S$ is set to be $H$-causing or $T$-causing.

$$\mathcal{J}_O = \{S \leftarrow H\text{-causing}, S \leftarrow T\text{-causing}\} \tag{9}$$

This note is long enough without considering interventions in detail. For the rest of this note, we will assume that the observer chooses not to intervene. This is equivalent to saying that their intervention distribution has all its mass on the null intervention which does not change the state of the referent:

$$\Pr_{\mathcal{J}_O \mid \mathcal{U}_O} = \{(U_O \leftarrow U_O, 1)\} \tag{10}$$

## 5 The Language Model

The definition of the term "language model" has blurred in recent years, and so deserves clarification. Here, we will consider a language model to be a conditional probability distribution $P$ over sequences of symbols $w$ (also called *tokens*) belonging to a vocabulary $\mathbb{V}$, paired with a sampling function $\textsc{Sample}$ : $P(\mathbb{V}) \times [0,1] \to \mathbb{V}$. Given an input sequence $w_1, ...w_l$, the model produces a probabilty distribution $P(w_{l+1} \mid w_1, ...w_l)$. To generate text, language models are combined with a sampling function that accepts a probability distribution and returns a next token. This sampling function may be deterministic, but it is often stochastic. To model stochastic sampling, we will assume the language model has a "source of randomness" like a pseudo-random number generator. Let us assume this generator returns a real value between 0 and 1 whenever it is called. Having externalized the random number generation, the $\textsc{Sample}$ function itself can be treated as a deterministic function[1]. It accepts the language model's next-token distribution and a random number $r \in [0,1]$, and returns the next token:

$$\textsc{Sample} : P(\mathbb{V}) \times [0,1] \to \mathbb{V} \tag{11}$$

Repeated application of the $\textsc{Sample}$ function allows us to generate an output sequence of multiple tokens. Thus, if the lengths of the input and output sequences ($l$ and $n$, respectively) are known, we can treat the behavior of the language model under repeated sampling as a deterministic function LM : $\mathbb{V}^l \to \mathbb{V}^n$, where $n$ is the length of the output sequence. The language model context size $c$ is a fixed value that dictates the maximum number of tokens the model can process. That is, $l + n \leq c$.

With sizes of the input and output sequences known, it is straightforward to view the language model as a causal model. Tokens $w_1, ..., w_n$ are the input tokens (exogenous variables), $r_{n+1}, ..., r_{n+l}$, $r_i \in [0,1]$ (also exogenous) are random numbers generated at each step, and $w_{n+1}, ..., w_{n+l}$ are the output tokens (endogenous variables).

$$\mathcal{U}_L = (w_1, ..., w_n, r_{n+1}, ..., r_{n+l}) \qquad \mathcal{V}_L = (w_{n+1}, ..., w_{n+l}) \tag{12}$$
$$\mathcal{R}_L(\mathcal{U}_L) = \mathbb{V}^n \times [0,1]^l \qquad \mathcal{R}_L(\mathcal{V}_L) = \mathbb{V}^l$$

---

[1]This covers widely-used sampling strategies like top-k and top-p. Accounting for sequence-level search strategies like beam search is more involved, this is skipped to expedite the example.

For notational convenience, let us partition $\mathcal{U}_L$ into $\mathcal{U}_L^w = (w_{n+1}, ..., w_{n+l})$ and $\mathcal{U}_L^r = (r_{n+1}, ..., r_{n+l})$. Variables $\mathcal{U}_L^w$ are the input tokens; these are provided by the observer, and their probability distribution is defined in Equation 1. Variables $\mathcal{U}_L^r$ are random numbers generated at each step, and their probability distribution is uniform on the interval $[0, 1]$.

$$P(r_i) = \text{Unif}(0, 1) \quad \text{for all } i \in [n + 1..n + l] \tag{13}$$

The language model generates tokens autoregressively, with each token depending on all of the input tokens, the preceding output token, and the random number generated at that step. This gives us a natural choice for the structural equations. For all $i \in [n + 1..n + l]$, we have a structural equation with signature $F_{w_i} : \mathcal{R}_L(\mathcal{U}_L^w) \times \mathcal{R}_L(\mathcal{U}_L^r)_i \times \mathcal{R}_L(\mathcal{V}_L)_{0..i-1} \to \mathcal{R}_L(w_i)$ and the following form:

$$F_{w_i} = \text{Sample}(P(w_i \mid w_1, ..., w_{i-1}), r_i) \tag{14}$$

For the sake of simplicity, this note is concerned only with single-turn interactions where the language model sees only a single input sequence from the observer. Additionally, the following examples are simplified by using input and output sequences of uniform length. Extensions to accommodate variable-length input and output and multi-turn interaction are included in Appendix Section 10.1.

## 6 The Observer's Mappings

Observers see simulacra when they can map between states of the simulator and states of the referent. This involves mapping exogenous states and endogenous states.

### 6.1 Mapping exogenous states

Let us imagine an observer who can think of only three ways to prompt a language model so that it behaves like a fair coin. For this rather unimaginative observer, these three prompts exhaust the ways to represent this referential scenario to the simulator. Let us imagine the observer is equally likely to use any of these prompts.

$$\begin{aligned}\Pr(\mathcal{U}_L \mid \mathcal{U}_O, \mathcal{I}_O = \text{null}) = \{&(H\text{-causing}, \text{"flip a coin"}) &\to 1/3, \\ &(H\text{-causing}, \text{"toss a coin"}) &\to 1/3, \\ &(H\text{-causing}, \text{"simulate a coin"}) &\to 1/3, \\ &(T\text{-causing}, \text{"flip a coin"}) &\to 1/3, \\ &(T\text{-causing}, \text{"toss a coin"}) &\to 1/3, \\ &(T\text{-causing}, \text{"simulate a coin"}) &\to 1/3\}\end{aligned} \tag{15}$$

A shorthand notation is used here and for the rest of this note to indicate states of the simulator (prompts to, or outputs from, the language model), where `"flip a coin"` means $\{w_1 = \text{"flip"}, w_2 = \text{"a"}, w_3 = \text{"coin"}\}$, `"Heads"` means $\{w_4 = \text{"Heads"}\}$, and so on.

### 6.2 Mapping endogenous states

Our observer also needs to decide which model-generated tokens to view as representing the event that the coin lands heads or tails. Let us say our observer accepts language model states $w_{n+1} = \text{Heads}$ as representing referent state $X = H$, and $w_{n+1} = \text{Tails}$ as representing referent state $X = T$.

$$\begin{aligned}\tau = \{&\text{Heads} \to H, \\ &\text{Tails} \to T\}\end{aligned} \tag{16}$$

Equations 5-10, 15, and 16 completely define an observer according to Definition 5.

# 7 Failure Cases

We can construct some cases in which the simulation fails, in the sense that the criterion for causal abstractive simulation (Definition 6) is not satisfied.

**Example 1. Failure due to sampling strategy** Consider a language model that has a nearly correct distribution over exogenous states relative to the observer's expectations (see $P_1$ below). The language model is poised to generate tokens that the observer maps to referent states (see $\tau$ above), and the probability distribution is close to what the observer has in mind. The language model's distribution is biased towards tokens that $O$ maps to $H$, but only slightly.

$$P_1 = \{... \\
(\text{"flip a coin"}, \quad \text{"Heads"}) \to 0.51, \\
(\text{"flip a coin"}, \quad \text{"Tails"}) \to 0.49, \\
(\text{"toss a coin"}, \quad \text{"Heads"}) \to 0.51, \\
(\text{"toss a coin"}, \quad \text{"Tails"}) \to 0.49, \\
(\text{"simulate a coin"}, \text{"Heads"}) \to 0.51, \\
(\text{"simulate a coin"}, \text{"Tails"}) \to 0.49, \\
...\} \tag{17}$$

The probability distribution above is not completely defined, but this is fine for our purposes; we only need to know the conditional probabilities that follow from prompts that the observer might use.

Consider the following two sampling functions:

$$\text{Sample}_{1.1} = \text{argmax}_{w_{l+1} \in \mathbb{V}} P(w_{l+1} \mid w_1, ... w_l) \tag{18}$$

$$\text{Sample}_{1.2} = \text{Top-2}\left(P(w_{l+1} \mid w_1, ... w_l), r_{l+1}\right) \tag{19}$$

$\text{Sample}_{1.1}$ is *greedy sampling*, a deterministic sampling function that returns the token with the highest probability at every generation step. $\text{Sample}_{1.2}$ is *top-2 sampling*, a non-deterministic sampling function that normalizes the largest two conditional probabilities from $P$. A definition for top-2 sampling is given in Appendix Section 10.2.

A language model with sampling function $\text{Sample}_{1.1}$ will fail to simulate the coin for $O$, beyond the gap expected from the biased distribution. Greedy sampling exacerbates the bias in the language model's probability distribution, such that the language model produces "Heads" for all of the observer's prompts. $\text{Sample}_{1.2}$ will fare much better.[2]

**Example 2. Failure due to language model probability distribution** We saw from Example 1 that greedy sampling is doomed to fail, and we would prefer a non-deterministic sampling strategy like Top-2. Here, we will see how a language model using Top-2 can fail to simulate, as a consequence of having the wrong probability distribution. Consider the following two conditional probability distributions:

---

[2]Empirical testing confirms this prediction: a language model $\langle P_1, \text{Sample}_{1.1}\rangle$ achieves a total variation distance of $0.500 \pm 0.000$. In contrast, the same model with $\text{Sample}_{1.2}$ achieves a dramatically lower total variation distance of $1.06 \times 10^{-2} \pm 3.0 \times 10^{-3}$. Total variation distance was measured between distributions $M_O(\text{Pr}_O)$ and $\tau_O(M_L(\text{Pr}_L))$ as described in Definition 7.

$P_{2.1} = \{...$
  ("flip a coin",  "Heads") $\to 0.9,$
  ("flip a coin",  "Tails") $\to 0.1,$
  ("toss a coin",  "Heads") $\to 0.9,$
  ("toss a coin",  "Tails") $\to 0.1,$
  ("simulate a coin", "Heads") $\to 0.9,$
  ("simulate a coin", "Tails") $\to 0.1$
 ...$\}$

$P_{2.2} = \{...$
  ("flip a coin",  "Heads") $\to 0.5,$
  ("flip a coin",  "Tails") $\to 0.5,$
  ("toss a coin",  "Heads") $\to 0.5,$
  ("toss a coin",  "Tails") $\to 0.5,$
  ("simulate a coin", "Heads") $\to 0.5,$
  ("simulate a coin", "Tails") $\to 0.5$
 ...$\}$

A language model $\langle P_{2.1}, \text{Top-2} \rangle$ will fail to simulate due to a strong bias towards tokens that $O$ maps to $H$. A language model $\langle P_{2.2}, \text{Top-2} \rangle$ will succeed.

**Example 3. Failure due to mismatch with observer's expectations** Examples 1 and 2 illustrated failure cases that can be "blamed on the language model", so to speak. The present example illustrates a case where the failure of simulation can arguably be blamed on the observer. Recall the observer's mapping from language model tokens to referent states $\tau$ in the preceding section. Assume the language model uses Top-2 sampling. Consider the following two conditional probability distributions:

$P_{3.1} = \{...$
  ("flip a coin",  "Heads") $\to 0.5,$
  ("flip a coin",  "Tails") $\to 0.5,$
  ("toss a coin",  "Heads") $\to 0.5,$
  ("toss a coin",  "Tails") $\to 0.5,$
  ("simulate a coin", "Heads") $\to 0.5,$
  ("simulate a coin", "Tails") $\to 0.5$
 ...$\}$

$P_{3.2} = \{...$
  ("flip a coin",  "H") $\to 0.5,$
  ("flip a coin",  "T") $\to 0.5,$
  ("toss a coin",  "H") $\to 0.5,$
  ("toss a coin",  "T") $\to 0.5,$
  ("simulate a coin", "H") $\to 0.5,$
  ("simulate a coin", "T") $\to 0.5$
 ...$\}$

A language model $\text{LM}_{3.1} = \langle P_{3.1}, \text{Top-2} \rangle$ will succeed, while $\text{LM}_{3.2} = \langle P_{3.2}, \text{Top-2} \rangle$ fails, since the tokens generated by $\text{LM}_{3.2}$ are not mapped to referent states by $\tau$. Of course, if the observer had a different mappping function, say $\tau'$ below, then both $\text{LM}_{3.1}$ and $\text{LM}_{3.2}$ would succeed.

$$\tau' = \{\text{"Heads"} \to H,$$
$$\text{"Tails"} \to T,$$
$$\text{"H"} \quad \to H,$$
$$\text{"T"} \quad \to T\} \tag{20}$$

## 8 A Successful Simulation
### Example 4. Successful simulation

This example combines elements of the prior examples to construct a case where the language model succeeds in simulating the coin for the observer.

Consider a language model $\text{LM}_4 = \langle P_4, \text{Top-2} \rangle$ where

$$P_4 = \{\ldots$$
$$(\text{"flip a coin"}, \quad \text{"Heads"}) \to 0.5,$$
$$(\text{"flip a coin"}, \quad \text{"Tails"}) \to 0.5,$$
$$(\text{"toss a coin"}, \quad \text{"Heads"}) \to 0.5,$$
$$(\text{"toss a coin"}, \quad \text{"Tails"}) \to 0.5, \quad (21)$$
$$(\text{"simulate a coin"}, \text{"Heads"}) \to 0.5,$$
$$(\text{"simulate a coin"}, \text{"Tails"}) \to 0.5$$
$$\ldots\}$$

We can show that $M_O(\text{Pr}_O)$ and $\tau_O(M_L(\text{Pr}_L))$ are identical. Let's calculate $M_O(\text{Pr}_O)$.

$$M_O(\text{Pr}_O) = M_O(\{(H\text{-causing}, 0.5), (T\text{-causing}, 0.5)\}) = \{H \to 0.5, T \to 0.5\} \quad (22)$$

Let's calculate $\text{Pr}_{\mathcal{U}_L^O}$, the distribution over observer-accessible exogenous variables of the simulator. Prompts are abbreviated here for legibility. The marginalization from Equation 1 is used to calculate $\text{Pr}_{\mathcal{U}_L^O}$; it is simplified here because the observer never intervenes (Equation 10).

$$\text{Pr}_{\mathcal{U}_L^O} = \sum_{\mathcal{U}_O} \sum_{\mathcal{I}_O} \text{Pr}_{\mathcal{U}_L^O \mid \mathcal{U}_O, \mathcal{I}_O} \text{Pr}_{\mathcal{I}_O \mid \mathcal{U}_O} \text{Pr}_{\mathcal{U}_O} \quad \text{Equation 1}$$

$$= \sum_{\mathcal{U}_O} \text{Pr}_{\mathcal{U}_L^O \mid \mathcal{U}_O, \mathcal{I}_O = \text{null}} \text{Pr}_{\mathcal{U}_O} \quad \text{observer never intervenes (Equation 10)}$$

$$= \{\text{"flip..."} \to P(\text{"flip..."}, HC)P(HC) + P(\text{"flip..."}, TC)P(TC),$$
$$\quad \text{"toss..."} \to P(\text{"toss..."}, HC)P(HC) + P(\text{"toss..."}, TC)P(TC), \quad (23)$$
$$\quad \text{"sim..."} \to P(\text{"sim..."}, HC)P(HC) + P(\text{"sim..."}, TC)P(TC)\}$$

$$= \{\text{"flip a coin"} \to 1/3,$$
$$\quad \text{"toss a coin"} \to 1/3,$$
$$\quad \text{"simulate a coin"} \to 1/3\}$$

,

We can now calculate $M_L(\text{Pr}_L))$, the distribution over endogenous variables of the simulator (language model outputs) resulting from the observer's inputs.

$$M_L(\text{Pr}_L) = \{\text{"Heads"} \to \quad P(\text{"Heads"}, \text{"flip..."})P(\text{"flip..."}) +$$
$$\quad P(\text{"Heads"}, \text{"toss..."})P(\text{"toss..."}) +$$
$$\quad P(\text{"Heads"}, \text{"sim..."})P(\text{"sim..."}),$$
$$\text{"Tails"} \to \quad P(\text{"Tails"}, \text{"flip..."})P(\text{"flip..."}) +$$
$$\quad P(\text{"Tails"}, \text{"toss..."})P(\text{"toss..."}) + \quad (24)$$
$$\quad P(\text{"Tails"}, \text{"sim..."})P(\text{"sim..."})\}$$
$$= \{\text{"Heads"} \to 1/2,$$
$$\quad \text{"Tails"} \to 1/2\}$$

Finally, we use $\tau$ to map the language model's outputs to referent states.

$$\tau_L(M_L(\Pr_L)) = \left\{ H \to \frac{1}{2}, \right.$$
$$\left. T \to \frac{1}{2} \right\} \tag{25}$$

We have shown that $M_O(\Pr_O) = \tau_L(M_L(\Pr_L))$, as both distributions assign probability $\frac{1}{2}$ to each outcome. According to Definition 6, the language model $\text{LM}_4$ successfully simulates the fair coin for observer $O$.

∎

## 9 Discussion

This note has illustrated several of the features of the theory of causal abstractive simulation. This theory proposes that simulations can be modeled using an ontology that includes observers, referents, simulators, and simulacra, and gives formal definitions for observers, referents, and simulators. This theory rests on the CASH: A simulator $S$ simulates a referent $R$ to an observer $O$ if $O$'s model of $R$ is a causal abstraction of $S$. This leads to a precise way of answering whether or to what extent a language model simulates some referent system to some observer, provided that the observer has a causal description of that referent system. Simulations can fail for a variety of reasons, some of which have been illustrated here. A success case was illustrated, as a basic example of how one might go about proving that a language model does indeed simulate another system.

# 10 Appendix

## 10.1 Extensions

**Variable-length output**     It is a minor inconvenience that the length of the output sequence is typically not known at the time that input is presented to the language model. Generation stops when the language model generates a STOP token, a stochastic event that may occur before $l + n$ reaches $c$. One way around this inconvenience is to assume that all output sequences are "padded" up to the size of the context window. That is, we treat an output sequence $s = (w_{n+1}, ..., w_{n+l})$ as $s\varepsilon^{c-(n+l)}$ where $\varepsilon^{c-(n+l)}$ is the empty token $\varepsilon$ repeated $c - (n + l)$ times. It might seem unrealistic to think that a typical observer will look at all of these empty tokens; recent models have $c \geq 1 \times 10^6$. As a more realistic alternative for human observers, we might pad up to some maximum output size dictated by the observer, replacing $c$ with some value $c_o < c$ in the padding scheme above. To enforce padding, the structural equations would be modified such that:

$$F_{w_i} = \begin{cases} \varepsilon & \text{if STOP} \in w_1, ..., w_{i-1} \\ \text{SAMPLE}(P(w_i \mid w_1, ..., w_{i-1}), r_i) & \text{otherwise} \end{cases} \quad (26)$$

Padding allows us to define the observer's $\tau$ as a function with a fixed number of arguments.

**Variable-length input**     Another inconvenience comes from the fact that users of language models do not always write prompts of equal length. Here also, padding is an option. This raises questions of whether we allow padding tokens to have any causal influence on the language model generation. That is, whether $P(w_i \mid w_1, ..., w_{i-1}) = P(w_i \mid w_1, ..., w_{i-1}, \varepsilon^k)$ in all cases. Neither case presents much difficulty for the formalism. Padding both input and output sequences is convenient since it gives the sets of endogenous and exogenous variables fixed sizes, and a fixed correspondence to positions in the input and output sequences.

**Multi-turn interaction**     The examples in this note focus on single-turn interaction, where an observer presents a single input sequence $s_1$ to the language model, and the model responds with a single output sequence $p_1$. Nothing about the language model prevents the observer from appending a response $s_2$ to the language model and presenting $s_1 p_1 s_2$ as a second prompt to the language model (a second "turn" in the dialogue), and considering the language model's response as part of the same simulation. Single-turn interaction is popular in certain benchmarking settings, but multi-turn interaction is how many of us use language model chatbots. One seemingly reasonable approach would decompose the multi-turn interaction $s_1, p_1, s_2, p_2$ into an interaction $s_1, p_1$ capturing the first turn, and a second single-turn interaction $s_1 p_1 s_2, p_2$ capturing the second turn ($s_1 p_1 s_2$ indicates the concatenation of $s_1$, $p_1$, and $s_2$). After this decomposition, simulation quality as determined by something like Definition 7 could be assessed for each single-turn interaction, giving a quality trajectory over the whole multi-turn interaction.

## 10.2 Top-2 sampling

**Definition 8. Top-2 sampling**
Top-2 sampling is a non-deterministic sampling function that samples a token from the two most probable tokens in $P(w_{l+1} \mid w_1, ...w_l)$. This is a specific form of the more recognizable Top-$k$ sampling, with $k = 2$. Let $w_{l+1}^1$ denote the most probable token in $P(w_{l+1} \mid w_1, ...w_l)$, and $w_{l+1}^2$ denote the second most probable. Let $p_1$ denote the conditional probability of $w_{l+1}^1$ and $p_2$ denote the conditional probability of $w_{l+1}^2$. The normalized conditional probability of $w_{l+1}^1$ is $p_{1'} = \frac{p_1}{p_1+p_2}$ and the normalized conditional probability of $w_{l+1}^2$ is $p_{2'} = \frac{p_2}{p_1+p_2}$. Then,

$$\text{Top-2}\left(P(w_{l+1} \mid w_1, ...w_l), r_{l+1}\right) = \begin{cases} w_{l+1}^1 & \text{if } r_{l+1} \leq p_{1'} \\ w_{l+1}^2 & \text{if } r_{l+1} \geq p_{1'} \end{cases} \qquad (27)$$